\def\year{2021}\relax
\documentclass[letterpaper]{article} 
\usepackage{aaai21}  
\usepackage{times}  
\usepackage{helvet} 
\usepackage{courier}  
\usepackage[hyphens]{url}  
\usepackage{graphicx} 
\usepackage{natbib}  
\usepackage{caption} 
\title{Domain Generalisation with Domain Augmented Supervised Contrastive Learning (Student Abstract)}
\author{
    Hoang Son Le, Rini Akmeliawati, Gustavo Carneiro
    \\
}
\affiliations{
    The University of Adelaide\\

    hoangson.le@student.adelaide.edu.au
}

\begin{document}

\maketitle

\begin{abstract}
    Domain generalisation (DG) methods address the problem of domain shift, when there is a mismatch between the distributions of training and target domains. Data augmentation approaches have emerged as a promising alternative for DG. However, data augmentation alone is not sufficient to achieve lower generalisation errors. This project proposes a new method that combines data augmentation and domain distance minimisation to address the problems associated with data augmentation and provide a guarantee on the learning performance, under an existing framework. Empirically, our method outperforms baseline results on DG benchmarks.   
\end{abstract}

\section{Introduction}
Domain shift is a classical machine learning problem, when the distribution of the training data that the model is calibrated on is different from the distribution of the target data that the model encounters when deployed \cite{quionero2009dataset} (a non-i.i.d problem). Domain generalisation (DG) is a domain shift problem that presumes access to training data from different but related domains, and that aims to construct a robust model that can make accurate predictions on latent target domains \cite{li2017deeper}.  


Data augmentation emerged as a promising alternative for DG, given that adversarial examples can be generated for training at negligible costs \cite{zhang2019unseen}.
However, data augmentation approaches provide no universal guarantee on prediction performance. More precisely, even in i.i.d settings, unsuitable data augmentation can severely impact the test performance.
For instance, for the MNIST dataset, (i) augmentations such as vertical/horizontal flipping and excessive rotation may generate unrealistic or even incorrect annotated adversarial examples, and (ii) using such augmented images for training may introduce the label noise issue and further complicate the problem.  

Methods to identify augmentation functions that improve test performances have been developed \cite{cubuk2019autoaugment}, which involve learning the magnitudes and application probabilities of a set of stochastic augmentation functions that give the best accuracy on the validation dataset. However, such methods are not applicable in a non-i.i.d setting, when the distribution of the validation dataset is not reflective of unseen target distributions. Moreover, for DG problems, data augmentation can violate the necessary conditions \cite{david2010impossibility}, by (1) increasing the source-target distance, and (2) violating the invariant labelling assumption.

\section{Methodology}
\begin{figure}[t]
  \includegraphics[width=\linewidth]{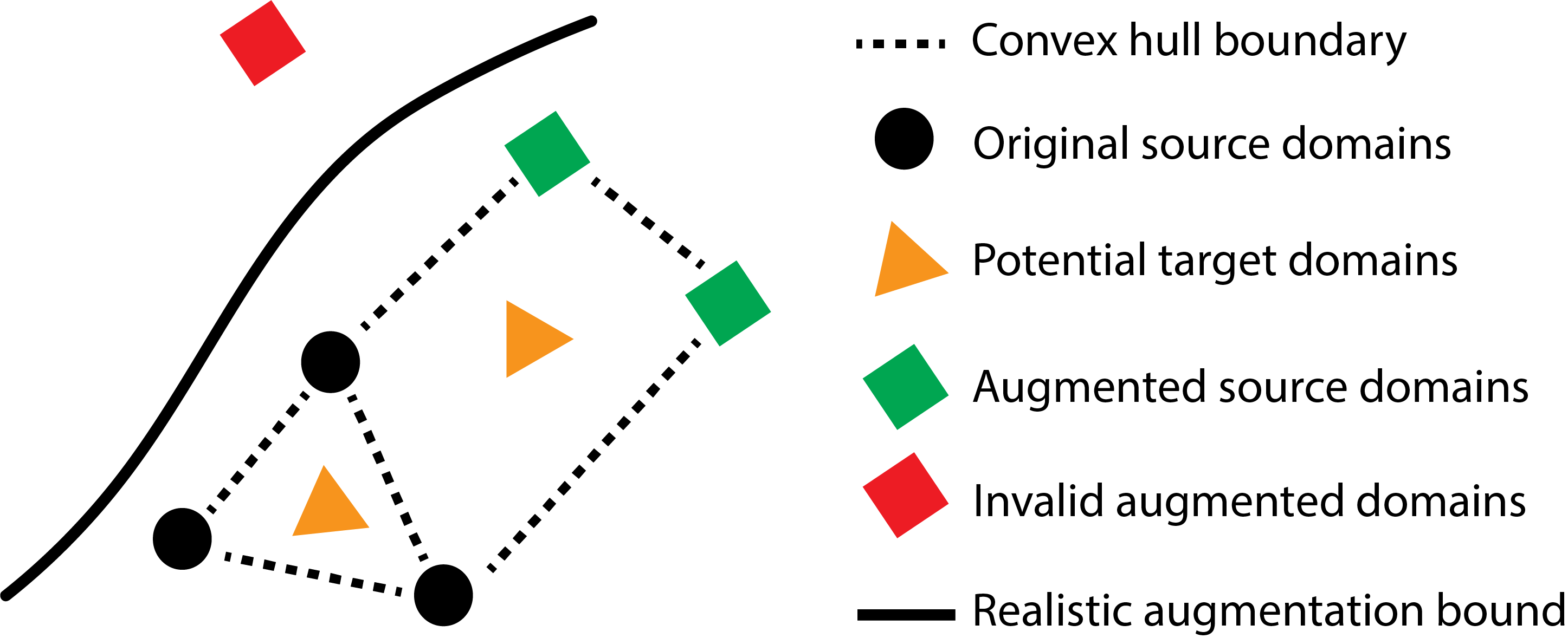}
  \caption{DASCL framework that combines realistic domain augmentation and domain distance minimisation.}
  \label{fig: Intuitive explanation}
\end{figure}

\citet{albuquerque2019generalizing} considered a special case of DG in which the target domain is a convex combination of the target domains. However, the number of target domains that this technique can cover is limited if the convex region of the training domains is small to begin with.  
Extending on this idea, we include target distributions that can be represented as a convex combination of the training and the augmented training domains (Figure \ref{fig: Intuitive explanation})

More specifically, we hypothesise a high dimensional space in which each axis is defined by an independent augmentation function, and the distance between two domains corresponds to the distortion magnitude of the combined augmentation functions. Under this framework, there is a distortion threshold for each augmentation, beyond which the invariant labelling assumption does not hold - i.e. (i) the augmented image is outside the domain of the labelling function, or (ii) the new label of the augmented image classified by the original labelling function is different from the original label. The set of all thresholds defines the feasible region that contains the original source domains and satisfies the invariant labelling assumption. From the theoretical framework in \citet{david2010impossibility}, the target distribution is constrained inside the feasible region.

Defining a realistic augmentation function is an ill-posed problem, and to the best of our knowledge, has not been explored. We propose a heuristic, based on the observation that it might be possible to generate realistic augmented domains that show a small reduction in validation accuracy, when a small augmentation magnitude is used. More concretely, we define an augmentation policy similar to \citet{cubuk2019autoaugment}, which consists of a set of augmentation functions, each having a probability of application and a transformation magnitude. The magnitudes are the highest values such that the resulting validation accuracy drop is still within a heuristic limit.

For each batch of training input, a composite augmentation function is drawn by sampling the augmentation policy, resulting in augmented domains whose distance from the original varies from zero (identity transformation) to the maximum. Based on the theory in \citet{albuquerque2019generalizing}, the target error bound can be improved by minimising the pair-wise distances between the training domains, which in this case, include the original and augmented domains. We propose to do this efficiently by using a supervised contrastive loss similar to \citet{khosla2020supervised}, which treats any pair of samples under the same label group as positive, irrespective of their input domains. This framework addresses some limitations of data augmentation in DG problem while improving the work in \citet{albuquerque2019generalizing} by considering a greater number of possible target domains that can be covered. 

\section{Results and Discussions}
We verify the performance of the proposed technique (DASCL) against the DG state-of-the-art (SOTA) - MMLD \cite{matsuura2019domain}, MASF \cite{dou2019domain}, and G2DM \cite{albuquerque2019generalizing} on the PACS \cite{li2017deeper} dataset and a multi-domain chest-x-ray medical dataset, consisting of Chexpert, Chest14, and Padchest \cite{irvin2019chexpert,bustos2020padchest,wang2017chestxray}. Overall, our method outperforms the standard ERM (deepall) benchmark on every domain, and yields the highest average improvement over all existing SOTA approaches. The empirical results yield an encouraging signal on our combined framework, given that the proposed data augmentation method is by no means optimal.

\begin{table}[t]

\centering
\scalebox{1}{
\begin{tabular}{c|c c c c |c} 
\hline
 Target &P & A  & C  & S  &Average \\ [0.1ex] 
\hline
MMLD    &88.98	 &69.27 &72.83 &66.44 &74.38\\
MASF    &\textbf{90.68}	 &70.35 &72.46 &67.33 &75.21\\
G2DM    &88.12   &66.60 &\textbf{73.3}6 &66.19 &73.55\\
Deepall &88.89   &68.14 &70.19 &61.07 &72.06\\
\hline
DASCL   &89.80 &\textbf{71.71} &71.55 &\textbf{72.77} &\textbf{76.41}\\
Std.dev &{0.81} &{1.17} &{1.14} &{0.98} &{0.39}
\end{tabular}
}
\caption{Mean and std dev accuracy results (in \%) on PACS using Alexnet. Targets are domains withheld from training, sources are the non-withheld domains used for training. The best results are highlighted in bold (up to the decimal value).}
\label{table:PACS_alexnet}

\end{table}

\begin{table}[t]

\centering
\scalebox{0.9}{
\begin{tabular}{c|c c c |c} 
\hline
 Target &Chexpert &Chest14  &Padchest &Average \\ [0.1ex] 
\hline
Deepall &76.53    &85.88  &83.13  &81.85\\
Std.dev &1.22     &2.28   &1.35   &0.93 \\
\hline
MMLD    &75.40 	  &87.70  &84.99  &82.70\\
Std.dev &2.30     &1.53   &0.29   &1.14 \\
\hline
DASCL   &\textbf{77.54}  &\textbf{88.83} &\textbf{87.31} &\textbf{84.55}\\
Std.dev &1.15     &0.92   &1.21   &0.62 \\
\end{tabular}
}
\caption{Mean and standard deviation of the classification AUC results (in \%) on the Medical dataset using Alexnet.}
\label{table:Medical_alexnet}
\end{table}

\addcontentsline{toc}{section}{References}
\bibliography{References}

\begin{thebibliography}{12}
\providecommand{\natexlab}[1]{#1}
\providecommand{\url}[1]{\texttt{#1}}
\providecommand{\urlprefix}{URL }
\expandafter\ifx\csname urlstyle\endcsname\relax
  \providecommand{\doi}[1]{doi:\discretionary{}{}{}#1}\else
  \providecommand{\doi}{doi:\discretionary{}{}{}\begingroup
  \urlstyle{rm}\Url}\fi

\bibitem[{Albuquerque et~al.(2019)Albuquerque, Monteiro, Darvishi, Falk, and
  Mitliagkas}]{albuquerque2019generalizing}
Albuquerque, I.; Monteiro, J.; Darvishi, M.; Falk, T.~H.; and Mitliagkas, I.
  2019.
\newblock Generalizing to unseen domains via distribution matching.

\bibitem[{Bustos et~al.(2020)Bustos, Pertusa, Salinas, and de~la
  Iglesia-Vay{\'a}}]{bustos2020padchest}
Bustos, A.; Pertusa, A.; Salinas, J.-M.; and de~la Iglesia-Vay{\'a}, M. 2020.
\newblock Padchest: A large chest x-ray image dataset with multi-label
  annotated reports.
\newblock \emph{Medical Image Analysis} 101797.

\bibitem[{Cubuk et~al.(2019)Cubuk, Zoph, Mane, Vasudevan, and
  Le}]{cubuk2019autoaugment}
Cubuk, E.~D.; Zoph, B.; Mane, D.; Vasudevan, V.; and Le, Q.~V. 2019.
\newblock Autoaugment: Learning augmentation strategies from data.
\newblock In \emph{CVPR 2019}, 113--123.

\bibitem[{David et~al.(2010)David, Lu, Luu, and
  P{\'a}l}]{david2010impossibility}
David, S.~B.; Lu, T.; Luu, T.; and P{\'a}l, D. 2010.
\newblock Impossibility theorems for domain adaptation.
\newblock In \emph{AISTAT 2010}, 129--136.

\bibitem[{Dou et~al.(2019)Dou, de~Castro, Kamnitsas, and
  Glocker}]{dou2019domain}
Dou, Q.; de~Castro, D.~C.; Kamnitsas, K.; and Glocker, B. 2019.
\newblock Domain generalization via model-agnostic learning of semantic
  features.
\newblock In \emph{NIPS 2019}, 6447--6458.

\bibitem[{Irvin et~al.(2019)Irvin, Rajpurkar, Ko, Yu, Ciurea-Ilcus, Chute,
  Marklund, Haghgoo, Ball, Shpanskaya et~al.}]{irvin2019chexpert}
Irvin, J.; Rajpurkar, P.; Ko, M.; Yu, Y.; Ciurea-Ilcus, S.; Chute, C.;
  Marklund, H.; Haghgoo, B.; Ball, R.; Shpanskaya, K.; et~al. 2019.
\newblock Chexpert: A large chest radiograph dataset with uncertainty labels
  and expert comparison.
\newblock In \emph{AAAI 2019}, volume~33, 590--597.

\bibitem[{Khosla et~al.(2020)Khosla, Teterwak, Wang, Sarna, Tian, Isola,
  Maschinot, Liu, and Krishnan}]{khosla2020supervised}
Khosla, P.; Teterwak, P.; Wang, C.; Sarna, A.; Tian, Y.; Isola, P.; Maschinot,
  A.; Liu, C.; and Krishnan, D. 2020.
\newblock Supervised Contrastive Learning.
\newblock \emph{arXiv preprint arXiv:2004.11362} .

\bibitem[{Li et~al.(2017)Li, Yang, Song, and Hospedales}]{li2017deeper}
Li, D.; Yang, Y.; Song, Y.-Z.; and Hospedales, T.~M. 2017.
\newblock Deeper, broader and artier domain generalization.
\newblock In \emph{ICCV}, 5542--5550.

\bibitem[{Matsuura and Harada(2019)}]{matsuura2019domain}
Matsuura, T.; and Harada, T. 2019.
\newblock Domain Generalization Using a Mixture of Multiple Latent Domains.
\newblock \emph{arXiv preprint arXiv:1911.07661} .

\bibitem[{Quionero-Candela et~al.(2009)Quionero-Candela, Sugiyama,
  Schwaighofer, and Lawrence}]{quionero2009dataset}
Quionero-Candela, J.; Sugiyama, M.; Schwaighofer, A.; and Lawrence, N.~D. 2009.
\newblock \emph{Dataset shift in machine learning}.
\newblock The MIT Press.

\bibitem[{Wang et~al.(2017)Wang, Peng, Lu, Lu, Bagheri, and
  Summers}]{wang2017chestxray}
Wang, X.; Peng, Y.; Lu, L.; Lu, Z.; Bagheri, M.; and Summers, R. 2017.
\newblock ChestX-ray8: Hospital-scale Chest X-ray Database and Benchmarks on
  Weakly-Supervised Classification and Localization of Common Thorax Diseases.
\newblock In \emph{CVPR 2017}, 3462--3471.

\bibitem[{Zhang et~al.(2019)Zhang, Wang, Yang, Sanford, Harmon, Turkbey, Roth,
  Myronenko, Xu, and Xu}]{zhang2019unseen}
Zhang, L.; Wang, X.; Yang, D.; Sanford, T.; Harmon, S.; Turkbey, B.; Roth, H.;
  Myronenko, A.; Xu, D.; and Xu, Z. 2019.
\newblock When unseen domain generalization is unnecessary? rethinking data
  augmentation.
\newblock \emph{arXiv preprint arXiv:1906.03347} .

\end{thebibliography}

\end{document}